\documentclass[sigconf]{acmart}
\usepackage{booktabs} 

\newcommand{\method}{\texttt{HeadArtist}\xspace}

\newcommand{\tocite}[1]{{\color{red} [TO CITE]}}

\usepackage{algorithmic}

\usepackage[ruled]{algorithm2e} 

\newcommand{\alliu}{\color{red}}

\usepackage[ruled]{algorithm2e} 

\SetAlFnt{\small}
\SetAlCapFnt{\small}
\SetAlCapNameFnt{\small}
\SetAlCapHSkip{0pt}
\AtBeginDocument{%
  }

\copyrightyear{2024}
\acmYear{2024}
\setcopyright{acmlicensed}\acmConference[SIGGRAPH Conference Papers '24]{Special Interest Group on Computer Graphics and Interactive Techniques Conference Conference Papers '24}{July 27-August 1, 2024}{Denver, CO, USA}
\acmBooktitle{Special Interest Group on Computer Graphics and Interactive Techniques Conference Conference Papers '24 (SIGGRAPH Conference Papers '24), July 27-August 1, 2024, Denver, CO, USA}
\acmDOI{10.1145/3641519.3657512}
\acmISBN{979-8-4007-0525-0/24/07}

\begin{document}

\title{HeadArtist: Text-conditioned 3D Head Generation with \\ Self Score Distillation}

\author{Hongyu Liu}
\authornote{This work is done partially when Hongyu is an intern at Ant Group.}
\email{hliudq@cse.ust.hk}
\affiliation{%
  \institution{HKUST}
  \city{Hong Kong}
  \orcid{0000-0002-4628-6388}
  \country{China}
}

\author{Xuang Wang}
\authornote{Joint corresponding authors.}
\affiliation{%
  \institution{Ant Group}
  \city{Hang Zhou}
  \orcid{0000-0001-5813-3875}
  \country{China}}

\author{Ziyu Wan}
\affiliation{%
  \institution{City University of Hong Kong}
  \orcid{0000-0003-2421-3484}
  \city{Hong Kong}
  \country{China}
}

\author{Yujun Shen}
\affiliation{%
 \institution{Ant Group}
 \city{Hang Zhou}
 \orcid{0000-0003-3801-6705}
 \country{China}}

\author{Yibing Song}
\affiliation{%
  \institution{Alibaba DAMO Academy}
  \city{Hang Zhou}
  \orcid{0000-0003-3667-531X}
  \country{China}}

\author{Jing Liao}
\affiliation{%
  \institution{City University of Hong Kong}
  \city{Hong Kong}
  \orcid{0000-0001-7014-5377}
  \country{China}}

\author{Qifeng Chen}
\authornotemark[2]
\affiliation{%
  \institution{HKUST}
  \city{Hong Kong}
  \orcid{0000-0003-2199-3948}
  \country{China}}
\email{cqf@ust.hk}


\begin{abstract}

We present \method for 3D head generation following human-language descriptions.
With a landmark-guided ControlNet serving as a generative prior, we come up with an efficient pipeline that optimizes a parameterized 3D head model under the supervision of the prior distillation itself. 
We call such a process \textbf{self score distillation (SSD)}.
In detail, given a sampled camera pose, we first render an image and its corresponding landmarks from the head model, and add some particular level of noise onto the image.
The noisy image, landmarks, and text condition are then fed into a frozen ControlNet \textbf{twice} for noise prediction.
 We conduct two predictions via the same ControlNet structure but with different classifier-free guidance (CFG) weights. The difference between these two predicted results directs how the rendered image can better match the text of interest.
%
Experimental results show that our approach produces high-quality 3D head sculptures with  rich geometry and photo-realistic appearance, which significantly outperforms state-of-the-art methods.
We also show that our pipeline supports editing operations on the generated heads, including both geometry deformation and appearance change. Project page:\url{https://kumapowerliu.github.io/HeadArtist}.

\end{abstract}


\begin{teaserfigure}
\includegraphics[width=1.\textwidth]{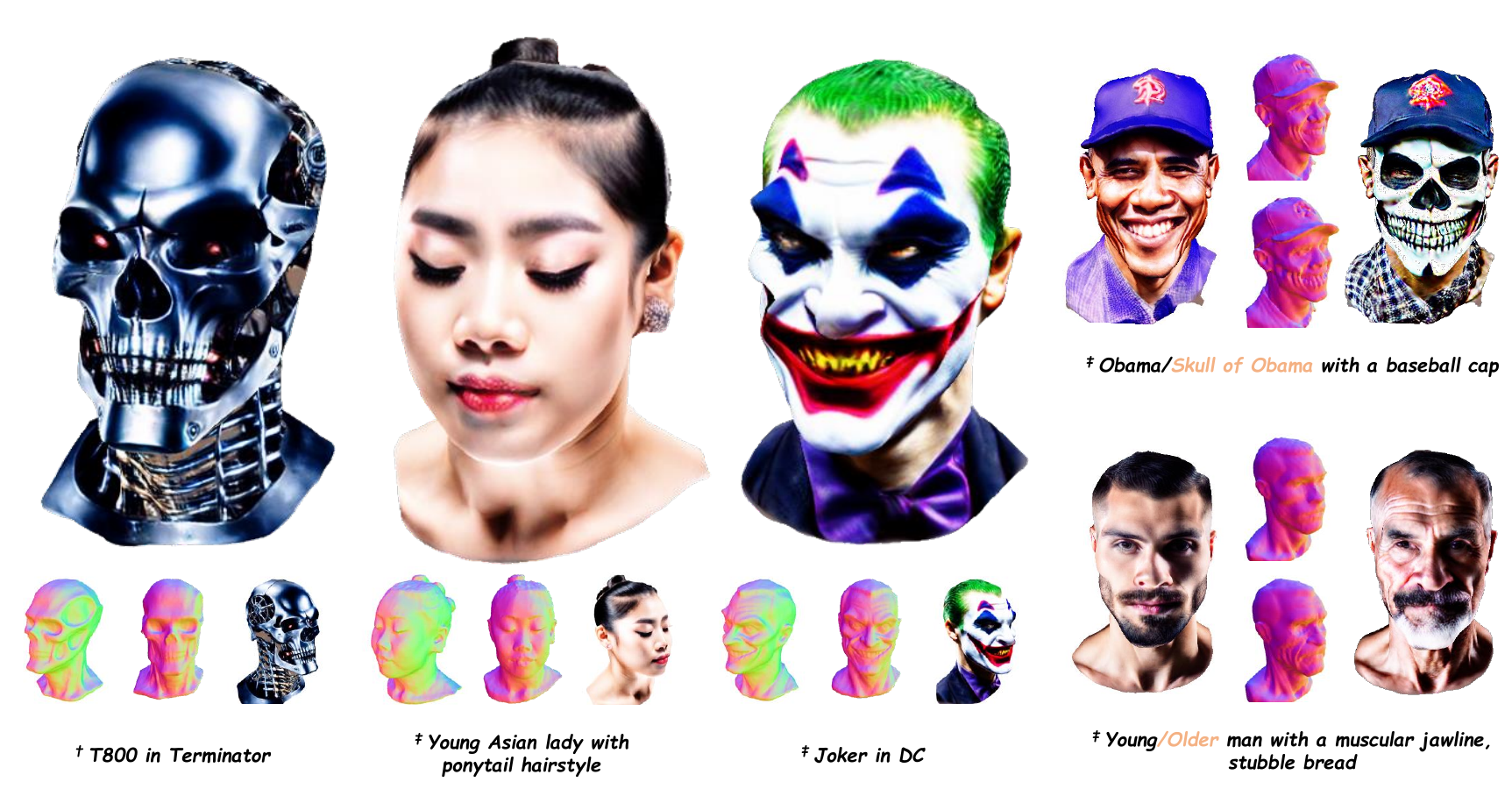}
\vspace{-2.5em}
    \caption{%
        \textbf{Generation and editing results} of our \method with text guidance, regarding both realistic and virtual characters.
        %
        %
        %
        $\dag$ and $\ddag$ denote the prefixes ``a head of ...'' and ``a DSLR portrait of ...'', respectively.
        Text in \textbf{\textcolor[rgb]{0.95,0.69,0.51}{orange}} denotes the editing instruction.
    }
\label{fig:teaser}
\end{teaserfigure}

\maketitle

\section{Introduction}
\label{sec:intro}
Advancements in vision-language models have significantly propelled the development of AIGC via human language descriptions, including text-to-image~\cite{ramesh2021zero,rombach2022high,ramesh2022hierarchical,nichol2021glide,saharia2022photorealistic,yu2022scaling}, 
%
%
text-to-video~\cite{singer2022make,ho2022imagen,wu2023tune,esser2023structure}, and text-to-3D generation~\cite{wang2023prolificdreamer,poole2022dreamfusion,lin2023magic3d,wang2023score,metzer2023latent, mendiratta2023avatarstudio}. Among these applications, text-to-3D head generation receives tremendous attentions. As part of the text-to-3D generation, modeling 3D head avatars holds significant potential in various areas including augmented reality, virtual reality, game character creation, and real-time interaction.

 There are several investigations~\cite{zhang2023styleavatar3d, yu2023towards,wang2023rodin} to generate text-guided human heads. Usually, they train a 3D generative model (i.e., Diffusion model~\cite{ho2020denoising} or GAN~\cite{goodfellow2014generative}) by using datasets consisting of text-image pairs.  With implicit 3D representations residing in these datasets, human heads are effectively generated. However, the performance of these methods hinges on high-precision dataset acquisition, and the generated heads sometimes lack diversity.


On the other hand, certain methods~\cite{huang2023avatarfusion,han2023headsculpt,cao2023dreamavatar,zhang2023avatarverse,chen2023fantasia3d} employ pre-trained text-to-image diffusion models to generate 3D heads, thereby are not limited by the dataset constraint and their results are more diverse. Specifically, they utilize Score Distillation Sampling~\cite{poole2022dreamfusion} (SDS) to optimize 3D head parameters with a pre-trained diffusion model. Facial landmarks~\cite{lugaresi2019mediapipe} or 3D Morphable Models~\cite{blanz2023morphable, FLAME:SiggraphAsia2017} (3DMM) are incorporated to enhance the generated outputs. However, these approaches cannot avoid limitations inherent to SDS, such as over-saturation and over-smoothing. Recently, Variational Score Distillation~\cite{wang2023prolificdreamer} (VSD) treats  3D parameters as random variables instead of constant data points as that in SDS. In this way, VSD can effectively address texture issues from SDS by utilizing normal CFG weights~\cite{ho2022classifier} (i.e., $\omega=7.5$). However, VSD may not produce satisfactory results when optimizing geometry for 3D head generation.  This is because VSD does not introduce a geometric structure prior of human heads (e.g., facial component). Moreover, both SDS and VSD may produce multi-face Janus artifacts on the 3D human heads. These artifacts reduce the fidelity and authenticity of the generated results.

In this paper, we present HeadArtist to distill 3D heads within a 3D head-aware diffusion model, by leveraging a frozen landmark-guided ControlNet~\cite{zhang2023adding}. Specifically, our method disentangles the generation process into geometry and texture generations. Besides, we utilize the DMTet mesh~\cite{shen2021deep} initialized with the Flame model~\cite{FLAME:SiggraphAsia2017} as our 3D head representation. Given a camera pose, we render an image and project the corresponding landmarks from the 3D head on it. We add noise to this rendered image, and combine this image with landmarks and text prompts to constitute our model inputs. These inputs are then fed into the ControlNet twice, enabling us to predict two types of noises. The first type of noise which is predicted by setting CFG as 1, represents the distribution score of the generated 3D head. We set another type of noise that is predicted by setting CFG as 7.5 and 100 for texture and geometry, respectively. This type of noise represents the distribution score of the target 3D head.  Our objective is to minimize the score difference between these two distributions for 3D head parameter optimizations. We name this process ``self-score distillation''. It represents the self-supervision of ControlNet on our 3D generation model. Intuitively, our SSD has two advantages: (1) As both of these two noises are sampled from the same ControlNet and landmarks, there is a direct spatial alignment of these two scores. This alignment effectively suppresses the occurrence of multi-face Janus artifacts to help us generate high-quality geometry; (2) As the landmarks already contain good facial structure priors, we use the ControlNet to naturally incorporate a 3D head prior during our generation process.

After generating a 3D head, we can manipulate both its geometry and texture by following human language descriptions. During manipulation and generation, direct use of negative text prompts can greatly enhance texture realism and intricacy. In Figure~\ref{fig:teaser}, we show that our head results are of premier quality from the perspective of geometry and texture. 

 
\begin{figure*}[t]
\begin{center}
\includegraphics[width=1.0\linewidth]{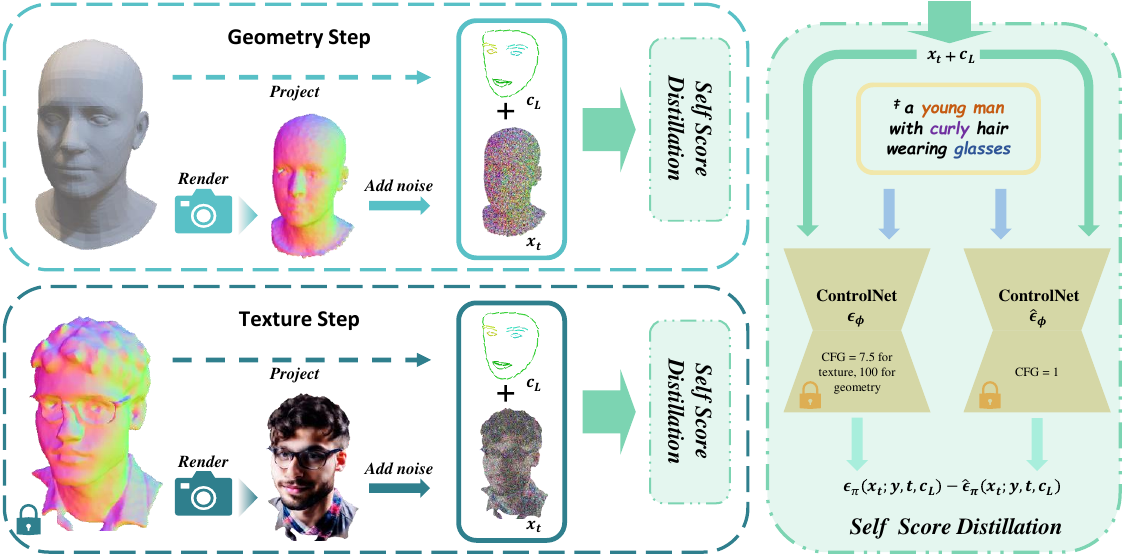}
\caption{We conduct two steps to generate geometry and textures, respectively. For the geometry step, we employ DMTet mesh~\cite{shen2021deep} initialized by Flame~\cite{FLAME:SiggraphAsia2017} as a representation of 3D human head's geometry. For the texture step, we fix the generated geometry and construct a texture space based on DMTet. The texture space construction is similar to that in Magic3D~\cite{lin2023magic3d}. In these two steps, we project mesh keypoints to get landmarks guided by a camera. Afterward, we render a normal map and a texture in these two steps for model training via our self-score distillation, the $x_t$ in these two steps are the normal map and texture with noise respectively. }
\vspace{-0.08in}
\label{fig:pipeline}
\end{center}
\end{figure*}

\section{Related Works}
In this section, we first discuss the progress of Text-to-3D generation on general objects. Then, we focus on the Avatar generation including human heads and human bodies.

\label{sec:relatedwork}
\subsection{Text-to-3D General Object Generation}
There has been a remarkable advancement in text-to-2D image generation~\cite{ramesh2021zero,rombach2022high,nichol2021glide,ramesh2022hierarchical,saharia2022photorealistic,yu2022scaling,zhu2022one}. These methods advance the text-to-3D content generation
~\cite{wang2023prolificdreamer, poole2022dreamfusion, lin2023magic3d,sanghi2022clip,wang2022clip}. Currently, text-to-3D methods generally use pre-trained text-to-2D image models to guide the generation process. One type of approaches~\cite{sanghi2022clip,wang2022clip,jain2022zero,michel2022text2mesh} adopt the  CLIP~\cite{radford2021learning} loss as supervision for model training. Due to the lack of spatial representation in CLIP, these methods contain challenges to producing authentic content. On the other hand, there is another type of approaches~\cite{lin2023magic3d,metzer2023latent,tang2023dreamgaussian,liu2023syncdreamer,sun2023dreamcraft3d,qian2023magic123,qiu2023richdreamer,liang2023luciddreamer} that leverage pre-trained diffusion models as supervision and utilize loss functions such as SDS~\cite{poole2022dreamfusion,wang2023score} to distill 3D parameters~\cite{mildenhall2021nerf, wang2021neus,shen2021deep}. While these approaches have achieved promising results, they suffer from the inherent problems of SDS, such as over-saturation, over-smoothing, and multi-face Janus artifacts. More recently, VSD~\cite{wang2023prolificdreamer} proposes a principled particle-based variational framework to generate realistic textures with normal classifier-free guidance weights. However, the VSD needs SDS to optimize the geometry, and it also contains multi-face Janus artifacts. 

\subsection{Text-to-3D Avatar Generation}

Apart from text-to-3D general object generation, several methods focus on text-to-3D Avatar Generation. Studies~\cite{zhang2023styleavatar3d,wang2023rodin} adopt the diffusion model~\cite{ho2020denoising} and generative adversarial networks~\cite{goodfellow2014generative} to produce the 3D head avatar. DreamFace~\cite{zhang2023dreamface} utilizes a component selection within the CLIP to generate preliminary geometry and employs SDS with a diffusion model to refine UV texture. Despite the impressive performance, these methods heavily rely on well-annotated datasets.  These datasets are challenging to acquire and constrain the generation diversity.  

Differently, several methods leverage pre-trained text-to-image models for supervising the training process without data dependence. Specifically, some methods~\cite{hong2022avatarclip, aneja2023clipface} use the CLIP loss~\cite{radford2021learning} to supervise the training process. They adopt the SMPL-X~\cite{pavlakos2019expressive} and Flame~\cite{FLAME:SiggraphAsia2017} as the geometry prior, respectively. These methods effectively generate corresponding avatars according to different texts and in different styles. DreamAvatar~\cite{cao2023dreamavatar}, DreamHuman~\cite{kolotouros2023dreamhuman} and Avatarfusion~\cite{huang2023avatarfusion} distillate a 3D avatar from a pre-trained 2D text-to-image diffusion model with SDS loss. They also apply some existing geometry parametric models to initialize the avatar shape. Moreover, a 3D aware diffusion model is trained in~\cite{zeng2023avatarbooth, huang2023humannorm, zhang2023avatarverse} to replace the pre-trained 2D diffusion model for achieving better performance. Meanwhile, some methods~\cite{shi2023mvdream, huang2023tech,liu2023zero,zhang2023text,yuan2023make, wu2023high} use an image accompanied with the text as guidance to increase model robustness. More recently, Head Sculpt~\cite{han2023headsculpt} introduces the prior-driven score distillation with landmark-guided ControlNet to improve 3D-head generations. Although these methods leverage SDS to generate realistic avatars, they cannot avoid intrinsic limitations associated with SDS. In this paper, we propose self-score distillation (SSD) to generate both geometry and texture for a substantial performance improvement.

\section{Preliminary}
In this section, we provide a brief overview of some essential prerequisites that are closely related to our HeadArtist in Section~\ref{sec:method}. We denote $\boldsymbol{\epsilon}_\phi$ as a pretrained text-to-2D diffusion model, $\theta$ as the parameters of a 3D representation (i.e., NeRF), $R$ as a differentiable rendering function and $y$ as the text prompt.

\label{sec:Preliminary}
\subsection{Score Distillation Sampling}
DreamFusion~\cite{poole2022dreamfusion} first proposes Score Distillation Sampling (SDS) that utilizes a pretrained text-to-2D diffusion model $\boldsymbol{\epsilon}_\phi$ to distill $\theta$. Specifically, for a rendered image $x=R(\theta, c)$ from a 3D model with a camera $c$, they first add a Gaussian noise ${\epsilon \sim \mathcal{N}(0,1)}$ to this image to get $x_t$ at time $t$ during the forward diffusion process. Then, they simulate the training process of the diffusion model to minimize the distance between the added noise ${\epsilon}$ and the predicted noise by using the pre-trained diffusion model. The gradient of SDS is computed as:
\begin{equation}
\small
\nabla_\theta \mathcal{L}_{\mathrm{SDS}}(R(\theta)) =\mathbb{E}_{t, \boldsymbol{\epsilon}}\left[\omega(t)\left(\boldsymbol{\epsilon}_\phi \left(\boldsymbol{x}_t; y, t\right)-\boldsymbol{\epsilon}\right) \frac{\partial x}{\partial \theta}\right],
\label{EQ:SDSGradient}
\end{equation} where $\omega(t)$ is a time-dependent weighting function. In practice, SDS utilizes the classifier-free guidance~\cite{ho2022classifier} (CFG) to adjust the sampling direction, which slightly deviates from unconditional sampling, i.e., $\epsilon_\phi\left(\boldsymbol{x}_t; y, t\right)+ W(\epsilon_\phi\left(\boldsymbol{x}_t ; y, t\right)-\epsilon_\phi\left(\boldsymbol{x}_t; t, \varnothing\right)$), where $\varnothing$ represents the ``empty'' text prompt, and $W$ is the CFG weight. The results show that SDS significantly improves text-to-3D performance. However, SDS is prone to produce over-saturated and over-smoothed results as shown in DreamFusion~\cite{poole2022dreamfusion}. This is because SDS needs a large CFG weight (i.e., $W=100$) for stable training. Also, SDS results contain multi-face Janus artifacts.

\subsection{Variational Score Distillation}
Recently, ProlificDreamer~\cite{wang2023prolificdreamer} considers $\theta$ as a random variable rather than a single point in SDS and proposes Variational Score Distillation (VSD). The VSD can use the normal CFG guidance weight (i.e.,  $W=7.5$) to get the distribution of high-quality results during training. Specifically, given a target text prompt $y$, there is a rendering image distribution $q^\mu\left(\boldsymbol{x} | y\right)$ of 3D representations $\mu(\theta | y)$ conditioned by this prompt. VSD tries to align this distribution with a real image distribution of $t=0$ defined by the pretrained text-to-image diffusion model under KL divergence as follows:
\begin{equation}
\small
\mathcal{L}_{\mathrm{VSD}}=\min _\mu D_{\mathrm{KL}}\left(q^\mu\left(\boldsymbol{x} \mid y,c\right) \| p\left(\boldsymbol{x} \mid y\right)\right) .
\label{EQ:VSDKL}
\end{equation}
To solve Eq.~\ref{EQ:VSDKL}, VSD formulates this optimization as the combination of different diffusion distributions indexed by $t$. Then, VSD calculates the distance between the distribution score of real and rendered images at time $t$ as below:
\begin{equation}
\begin{aligned}
\small
\nabla_\theta \mathcal{L}_{\mathrm{VSD}}( R(\theta))=\mathbb{E}_{t, \boldsymbol{\epsilon}} [& \omega(t) (\boldsymbol{\epsilon}_\phi (\boldsymbol{x}_t ; y, t )- \\ 
& \boldsymbol{\epsilon}_{\mathrm{lora}} (\boldsymbol{x}_t ; y, t, c ) ) \frac{\partial \boldsymbol{x}}{\partial \theta} ], 
\label{EQ:VSDGradient}
\end{aligned}
\end{equation}
where  $\boldsymbol{\epsilon}_{\text {lora }}$, a LoRA (Low-rank adaptation)~\cite{hu2021lora} model based on $\boldsymbol{\epsilon}_\phi$, represents the distribution of the rendered image. The LoRA model is trained by using the rendered image $x$, text prompt $y$, and camera pose $c$ to minimize the difference   $\left\|\boldsymbol{\epsilon}_{\text {lora }}\left(x_t, y, t, c\right)-\boldsymbol{\epsilon}\right\|_2^2$. However, optimizing the geometry of  $\theta$  directly with VSD is challenging and requires the guidance of SDS.  Moreover, the training process of LoRA is unstable and ineffective in obtaining an accurate representation of the current 3D distribution. So more time is taken for model training to converge. In addition, generating the 3D head directly using VSD is difficult due to the absence of a 3D head structure prior. With these observations, our method tries to address these issues with self score distillation.

\subsection{3D Representation}
We follow Fantasia3D~\cite{chen2023fantasia3d} to disentangle the geometry and texture. For the geometry,  we use DMTet~\cite{shen2021deep} as our geometry representation. Specifically, DMTet maintains a deformable tetrahedral grid ($V_T, T$) where $V_T$ are vertices. For each vertex $v_i \in V_T$, DMTet predicts a Signed Distance Function (SDF) value and a position offset from a learned MLP module. Subsequently, the SDF can be transferred to a mesh with a differentiable Marching Tetrahedral layer.  With the DMTet mesh, we can efficiently render high-resolution textures for fast training with effective detail preservation.

\section{Proposed Method}
\label{sec:method}
Head-Artist generates high-fidelity 3D heads following language descriptions.
Figure~\ref{fig:pipeline} shows an overview where our framework is divided into two steps: geometry and texture optimization steps. In the texture optimization step, we fix the optimized mesh as the shape guidance. The whole optimization process is conducted by self score distillation. Overall, our HeadArtist generates diverse 3D heads ranging from real individuals to virtual characters. The generated 3D heads contain intricate geometry and high-fidelity textures. The detailed illustration is shown below:

\subsection{Self Score Distillation}

The general objects are generated well by SDS and VSD. For 3D human heads, these two methods encounter substantial challenges as shown in Figure~\ref{fig:compare}. This is because 3D heads contain two unique characteristics apart from general objects: 1) Unlike general objects such as hamburgers with viewpoint-invariant properties, 3D heads exhibit obvious view discrepancy across different poses. The view discrepancy makes the generation more susceptible to multi-face Janus artifacts; 2) The facial region of 3D heads contains highly structured semantic components (i.e., eyes, nose, and mouth). These components have been precisely identified. Therefore, the pose and facial components are important guidance and contribute to the 3D head generation process. Although HeadSculpt~\cite{han2023headsculpt}  leverages a 3D-head-aware diffusion model, it introduces SDS to generate 3D heads. As such, the inherent limitations of SDS (i.e., over-saturation and over-smoothing) will affect its output result.


We propose Self Score Distillation (SSD) to obtain a 3D head distribution within a pre-trained 3D head-aware diffusion model. Specifically, there is a distribution of a rendered image $q^{\mu_{h}}(\boldsymbol{x} | y, c)$  with optimized 3D head distribution $\mu_{h}(\theta | y)$. Also, we can obtain a pre-trained 3D-head-aware diffusion model to get a real image distribution $p(\boldsymbol{x}| y, c)$  with a text and camera pose. Finally, our self score distillation aims to minimize the following KL divergence: 
\begin{equation}
\small
\mathcal{L}_{\mathrm{SSD}}= \min_{\mu_{h}}D_{\mathrm{KL}}\left(q ^{\mu_{h}}\left(\boldsymbol{x} \mid y, c \right) \| p \left(\boldsymbol{x}  \mid y, c\right)\right) .
\label{EQ:SD}
\end{equation}

Inspired by VSD and SDS, to address the above KL divergence, we can set up a series of optimization objectives characterized by varying marginal diffused distributions indexed by $t$:
\begin{equation}
\small
   \mathcal{L}_{\mathrm{SSD}} =\underset{\mu_{h}}{\min }D_{\mathrm{KL}}(q_t^{\mu_{h}}(\boldsymbol{x}_t \mid y, t, c )  \|  p_t (\boldsymbol{x}_t \mid y, t, c  ) ) .
\label{EQ:SDKL}
\end{equation}
 
To solve Eq.~\ref{EQ:SDKL}, we need two 3D-head-aware diffusion models to represent the distribution of $q_t^{\mu_{h}}$ and $p_t$, respectively, so that we can sample   $x_t$ based on camera and text. For $p_t$, we employ a pre-trained landmark-guided ControlNet. This ControlNet is trained on a comprehensive 2D facial dataset~\cite{controlnet,zheng2022general}, where the ground-truth data is generated by rendering facial landmarks from MediaPipe~\cite{lugaresi2019mediapipe}. The incorporation of ControlNet allows us to obtain $x_t$ with multi-view consistency since we can denote the landmark as a camera pose. To compute  $q_t^{\mu_{h}}$, one possible approach is to train a LoRA model to make a prediction like what VSD does. However, the training process of LoRA is unstable.  Even though we assume LoRA has been trained well, the mismatch between LoRA and ControlNet makes the learned LoRA hard to align these two distributions. The mismatch is caused by these two diffusion models that adopt the conventional camera parameters and landmarks as head pose, respectively. Moreover, it leads to the emergence of the multi-face Janus artifact, as depicted in column (c) of Figure~\ref{fig:ablation}.

Fortunately, we observe that the ControlNet itself is trained on a meticulously aligned dataset consisting of facial landmarks, text, and face images. The accurate alignment enables ControlNet to naturally represent $q_t^{\mu_{h}}$ and sample $x_t$. The $x_t$ is the marginal distribution of the ControlNet given the text and landmarks. Different from LoRA, the ControlNet incorporates landmarks that provide significantly richer semantic information for human facial regions than only using camera parameters. The gradient of $\mathcal{L}_{\mathrm{SSD}}$ can be written as
\begin{equation}
\begin{aligned}
\small
\nabla_\theta \mathcal{L}_{\mathrm{SSD}}(  R(\theta))=\mathbb{E}_{t, \boldsymbol{\epsilon}} &[\omega(t) (\boldsymbol{\epsilon}_\pi (\boldsymbol{x}_t ; y, t, c_L )- \\ 
& \boldsymbol{\hat{\epsilon}}_{\pi} (\boldsymbol{x}_t ; y, t, c_L ) ) \frac{\partial \boldsymbol{x}}{\partial \theta} ], 
\label{EQ:SSDGradient}
\end{aligned}
\end{equation}
where $c_L$ is the landmark,  $\boldsymbol{\epsilon}_{\pi }$ and $\boldsymbol{\hat{\epsilon}}_{\pi}$ represent two pre-trained and frozen Controlnets with the same parameters. In practice, we employ  $\text{CFG}=1$ for $\boldsymbol{\hat{\epsilon}}_{\pi}$ to estimate the score of the optimized 3D head distribution based on the text. Meanwhile, we set different CFG weights for predicting the score of the real image distribution of geometry and texture. The detailed settings are illustrated in the next subsection. In general, we iteratively conduct the training process by using ControlNet with self-distillation. Two predicted noises are guided by the same landmarks, which ensures an accurate alignment.

In comparison to SDS and VSD, Our SSD offers two advantages: 1) we obtain two distribution scores from the same diffusion model via identical landmarks as shown in Eq.~\ref{EQ:SSDGradient}, which ensures accurate spatial alignment to address the multi-face Janus issue; 2) we introduce landmarks for 3D head generation. The landmarks benefit 3D heads as they naturally bring facial semantic features. By leveraging these two advantages, our method exhibits superior performance for 3D head generation.

\subsection{Head Generation}
We generate high-fidelity 3D heads via SSD training. Specifically, we first generate the geometry, then we fix the geometry as guidance to generate the texture. 

\subsubsection{Geometry Generation}

We follow Fantasia3D to adopt the DMTet as our geometry representation and initialize it with the Flame model. Specifically, the DMTet is parametrized via an MLP module $\psi_g$. For each vertex $v_i$ in a deformable tetrahedral grid ($V_T, T$), we use the MLP to predict SDF $s_{v_i}$ and offset $\Delta v_i$. We randomly sample a point set $\left\{p_i \in \mathbb{R}^3\right\}$ and get SDF value $\text{Flame}_{\text{SDF}}(p_i)$ in the Flame model, then we use the following loss to optimize $\psi_g$: $\sum_{p_i \in P} \|s (p_i )-\text{Flame}_{\text{SDF}}(p_i ) \|_2^2$. 

After initialization, given a camera, we use the differentiable render (e.g., nvidiffrast~\cite{Laine2020diffrast}) to generate the normal map $n$, and we project the specific vertices of Flame into a landmark map $c_L$. Then we add Gaussian noise to $n$ to get $x_t$ and use our SSD to optimize $\psi_g$ of  DMTet. Moreover, since there is a domain gap between the normal map and the real image which is used to train ControlNet, we set the  $W=100$ (CFG)  for ControNet $\boldsymbol{\epsilon}_{\pi }$ to stabilize training. Meanwhile, we render the opacity mask combined with the normal map in the early stages of optimization.

\subsubsection{Texture Generation}

During the texture generation process, the geometric structure remains fixed, while a neural color field is constructed via an MLP module. The MLP module is denoted as $\psi_{tex}$. The color field predicts the RGB value for each vertex in the geometry. Then, we render the color field to get the texture, and we add the noise to the texture to get the $x_t$. Meanwhile, we project the landmark with a fixed geometry structure. Finally, we send $x_t$, landmark, and text prompt to SSD to optimize the $\psi_{tex}$. For texture generation, we set  $W=7.5$ (CFG) for $\boldsymbol{\epsilon}_{\pi }$, which is the common setting for diffusion models to get high-fidelity results. Moreover, we have discovered that our SSD can effectively leverage the negative prompts (i.e., replacing the empty prompt in CFG with ``worst quality, low quality, semi-realistic") for further performance improvement.

\subsection{Head Editing}
 After high-quality 3D head generation, we can edit the generated head as a follow-up. The editing process contains two unique steps including geometric deformation and texture manipulation. Specifically, when users provide specific instructions to modify a generated head, we fix the SDF $s_{v_i}$ and focus on learning the offset $\Delta v_i$ for each vertex within DMTet to achieve mesh deformation.  Then, we fix the modified geometry, and continue to update parameters of $\psi_{tex}$ which is already pretrained in the generation process. The whole process is conducted by SSD with different text and landmarks. Since we establish a canonical space with fixed SDF values and the editing texture inherits from the previous generation, our final result can well preserve the identity of the original character.

\section{Experiments}

In this section, we illustrate our implementation details. Then, we provide qualitative and quantitative evaluations. Finally, we perform an ablation study of our approach.

\subsection{Implementation Details}
Our Head-Artist employs the Stable-Diffusion based ControlNet~\cite{controlnet} within the Huggingface Diffusers~\cite{von2022diffusers} with the version 2-1-base~\cite{sd21base}. The entire framework is constructed upon threestudio~\cite{threestudio2023}. For both geometry and texture generation, we render normals and textures at a size of 512$\times$512. We conduct 15k and 20k iterations for geometry and texture training, respectively. Our training is performed on a single NVIDIA RTX A6000 GPU, with a batch size of 1. The total training time is around 3 hours with an AdamW optimizer. We follow the ProlificDreamer~\cite{wang2023prolificdreamer} to utilize the annealed time schedule. During both texture and geometry generation, we follow the DreamFusion~\cite{poole2022dreamfusion} and add the view-dependent text for the ControNet $\boldsymbol{\epsilon}_{\pi }$. Specifically, for the back-view, we exclude facial features in the landmark and solely project the head's contour. We maintain a Hash Encoding~\cite{muller2022instant} to embed the poison points during the whole generation process. We set the maximum time steps as 0.5 for the geometry generation, since the geometry is initialized with the Flame model and need not generate from the complete Gaussian noise.  More implementation details and results can be found in the supplementary material.

We conducted comprehensive comparisons between our method and five approaches: DreamFusion~\cite{poole2022dreamfusion}, LatentNerf~\cite{metzer2023latent}, Fantasia3d~\cite{chen2023fantasia3d}, ProlificDreamer~\cite{wang2023prolificdreamer}, and HeadSculpt~\cite{han2023headsculpt}.  For fair comparisons, we incorporated Flame as the geometry prior for both LatentNerf and Fantasia3d. All methods except HeadSculpt are implemented with threestudio. As the implementation of HeadSculpt is not available, we directly borrow their provided results.

\subsection{Qualitative Evaluation}

 \begin{figure*}[t]
\begin{center}
\includegraphics[width=1.0\linewidth]{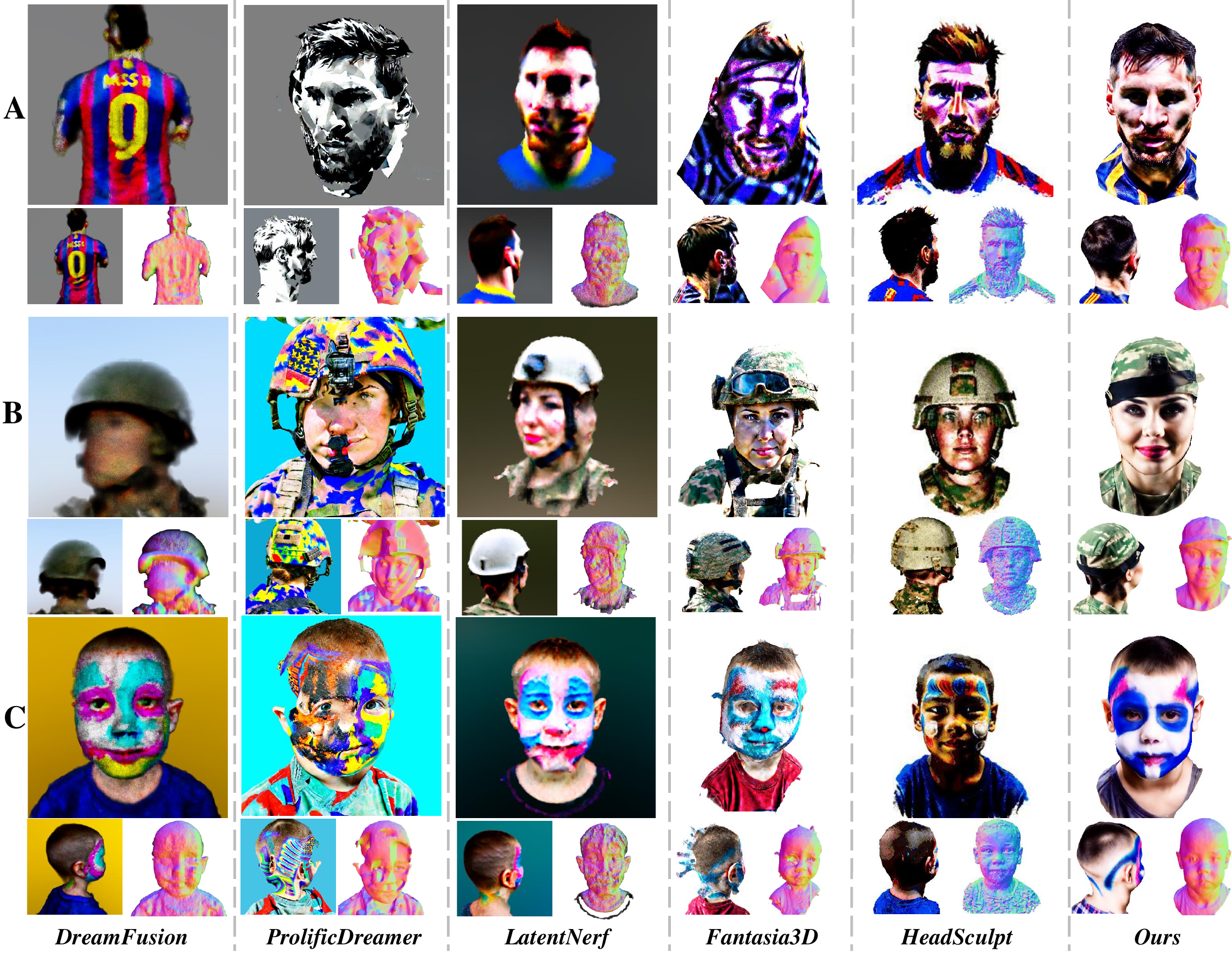}
\caption{Qualitative comparison with SOTA methods. Text Prompts: (A) a DSLR portrait of Lionel Messi. (B) a DSLR portrait of a female soldier, wearing a helmet. (C) a DSLR portrait of a boy with facial painting. Overall, our method demonstrates superior fidelity in terms of both geometry and textures.}
\vspace{-0.23in}
\label{fig:compare}
\end{center}
\end{figure*}

In Figure~\ref{fig:compare}, we present the visual comparisons between our HeadArtist and other methods.  When inspecting geometry,  LatentNerf, DreamFusion, and ProlificDreamer suffer from the problem of multi-face Janus. The HeadSculpt and Fantasia3d try to overcome this problem by introducing the 3D head prior, but their geometry still contains artifacts (e.g., the hollowness around the eyes in C of Fantasia3D, and the geometry noise of the female in B of HeadSculpt). In contrast, our HeadArtist, with the spatial alignments guided by landmarks, resolves the issue of multiple head artifacts and achieves more accurate head geometry.  When inspecting texture, due to the adoption of SDS during training, DreamFusion, LatentNerf, Fantasia3d, and HeadSculpt have either over-saturation or over-smooth issues. ProlificDreamer performs well in texture generation, but it has collapsed shapes with the multi-face Janus issue.  In comparison, our HeadArtist with SSD demonstrates more photorealistic texture generations. To sum up, our method generates human head geometries that are both plausible and free of artifacts, and the texture is remarkably realistic. Moreover, the generated results closely align with human-language descriptions. 

\begin{figure*}[t]
\begin{center}
\includegraphics[width=1.0\linewidth]{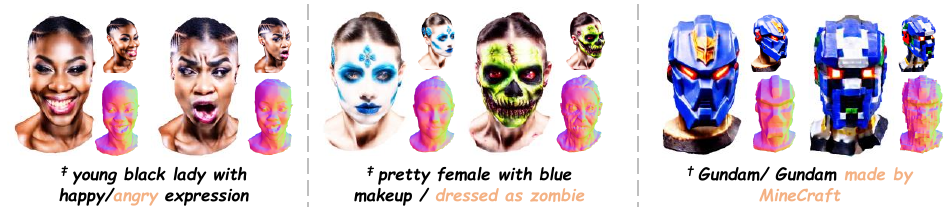}
\caption{The editing results of our HeadArtist. Our approach enables the manipulation of geometry and textures, thereby aligning the resulting content with the corresponding human-language descriptions. }
\vspace{-0.23in}
\label{fig:editing}
\end{center}
\end{figure*}

In addition to generation, our method is effective in editing. As depicted in Figure~\ref{fig:teaser} and Figure~\ref{fig:editing}, with text instructions, we can effectively make corresponding manipulations to the geometry and texture simultaneously. Our method enables the manipulation of various head attributes (i.e., expressions, age, and style), while effectively preserving the character identity.


\begin{figure*}[t]
\begin{center}
\includegraphics[width=1\linewidth]{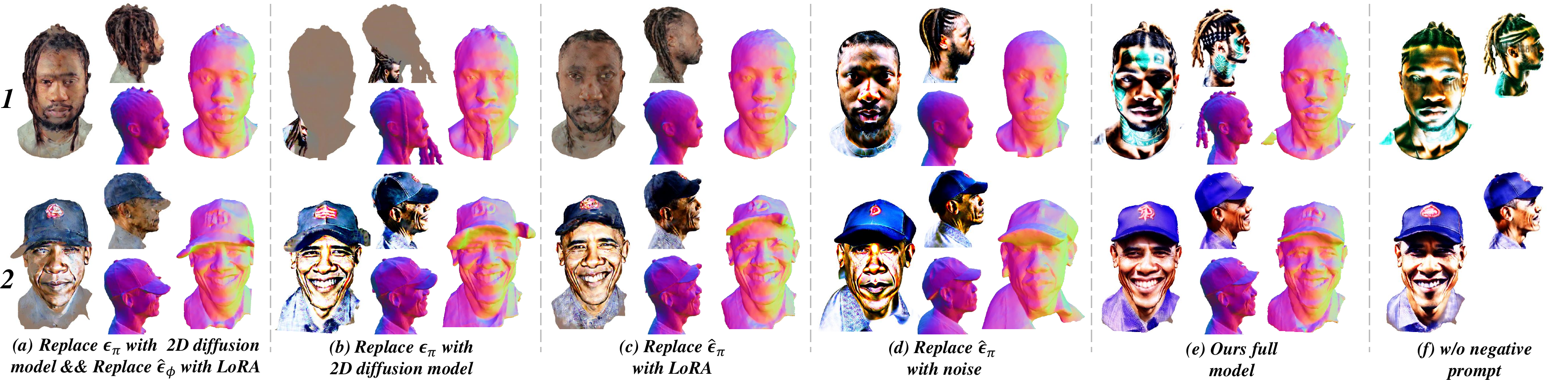}
\caption{Ablation study. Text Prompts: 1) a DSLR portrait of a young man with dreadlocks; 2) a DSLR portrait of Obama with a baseball cap. Compared with (a), (b), and (c), our full configuration in (e) avoids the multi-face Janus problems (e.g., multiple brims of Obama). Meanwhile, (e) can generate complex geometries which are better match the text(i.e., (a)$\sim$(d) can not generate the dreadlocks). Besides, (e) predicts more natural and fidelity colors and helps the texture fit the geometry better than (d). (f) shows our results without negative prompts.}
 \vspace{-0.19in}
\label{fig:ablation}
\end{center}
\end{figure*}

\subsection{Quantitative Evaluation}
We conduct a user study to evaluate the robustness of our method using 10 text prompts, which are from HeadSculpt.  Based on these text prompts, we compare the generated heads which are rendered into freely viewable videos showing both geometry and textures. We invite 10 participants with computer graphics backgrounds to rate each result with a score from 1 to 6 (the higher the score, the better the effect). We calculate the average score for each method. Table~\ref{tab:quantitative} shows the subjective evaluation performance where our method is more favorable.

\begin{table}[t]
\caption{%
    Quantitative comparisons of state-of-the-art methods. We conduct a user study to score the performance of the method, with a maximum score of 6 points. ${\uparrow}$ indicates higher is better.
}
\vspace{-8pt}
\centering
\setlength{\tabcolsep}{5pt}
\resizebox{1\linewidth}{!}{
\begin{tabular}{c|lclcccc}
\toprule
Method                                                                &  & CLIP-L/14 $\uparrow$                       &  & CLIP-B/32 $\uparrow$                              &  & User Study  $\uparrow$    &  \\
\midrule
DreamFusion~\cite{poole2022dreamfusion}                              &  & 0.261                                      &  & 0.284                                            &  & 2.06                      &  \\
\multicolumn{1}{l|}{ProlificDreamer~\cite{wang2023prolificdreamer}}  &  & 0.264                                      &  & 0.316                                             &  &  2.45                     &  \\
LatentNerf~\cite{metzer2023latent}                                   &  & 0.262                                      &  & 0.313                                             &  & 2.72                      &  \\
Fantasia3d~\cite{chen2023fantasia3d}                                 &  & 0.271                                      &  & 0.312                                             &  & 3.47                      &  \\
HeadSculpt~\cite{han2023headsculpt}                                  &  & 0.280                                      &  & 0.323                                             &  & 4.56                      &  \\
\textbf{Ours}                                                        &  & \textbf{0.300}                             &  & \textbf{0.334}                                            &  & \textbf{5.67}             &  \\
\bottomrule
\end{tabular}

}

\label{tab:quantitative}
\vspace{-6pt}
\end{table}

In addition to the user study, we further evaluate the relevance of text prompts and 3D heads by using the CLIP score~\cite{radford2021learning}. We use two CLIP models, including CLIP-B/32 and CLIP-L/14, to do the evaluation. We convert videos used in the User Study into images, and then feed these images, along with their corresponding texts, into the CLIP model to compute the scores. The results, as shown in Table~\ref{tab:quantitative}, indicate that our method is capable of generating 3D head models that exhibit a better alignment with the text prompts.
 
\subsection{Ablation Study} 

\subsubsection{ Effectiveness of SSD}
\begin{table}[t]
\caption{%
Quantitative evaluation of ablation study. The methods correspond one-to-one to the methods in Figure~\ref{fig:ablation}. ${\uparrow}$ indicates higher is better.
}
\vspace{-8pt}
\centering
\setlength{\tabcolsep}{5pt}
\resizebox{0.6\linewidth}{!}{
\begin{tabular}{c|lclcc}
\toprule
Method                               &  & CLIP-L/14 $\uparrow$  &  & CLIP-B/32 $\uparrow$    &           \\
\midrule
a                                    &  & 0.2590                &  & 0.2786                       &           \\
b                                    &  & 0.2132                 &  &  0.2351                      &           \\
c                                    &  & 0.2655               &  & 0.2901                          &           \\
d                                    &  & 0.2797                &  & 0.2986                           &           \\
f                                    &  & 0.2893                 &  & 0.3191                          &           \\
\textbf{e}                        &  & \textbf{0.2998}     &  & \textbf{0.3276} & \\
\bottomrule
\end{tabular}
\vspace{-20pt}
}

\label{tab:ablation}
\end{table}
We conduct different configurations in Figure~\ref{fig:ablation} to evaluate our SSD. In (a), we replace $\boldsymbol{ {\epsilon}}_{\pi}$ and $\boldsymbol{\hat{\epsilon}}_{\pi}$ with a 2D text-to-image diffusion model and LoRA model, respectively, which follows VSD. In (b), we replace $\boldsymbol{ {\epsilon}}_{\pi}$  with a 2D text-to-image diffusion model. The results in (a) and (b) show that multi-face Janus artifacts occur (i.e., two brims with Obama) and poor performance for texture and geometry coherence. These two configurations can be regarded as distilling the 3D head with a 2D diffusion model. As the 2D diffusion model lacks prior knowledge of heads and cannot be guided by the camera, the performance is limited. In (c), we change $\boldsymbol{\hat{\epsilon}}_{\pi}$ to LoRA. In (d), we replace $\boldsymbol{\hat{\epsilon}}_{\pi}$ by noise following SDS, and we set CFG of $\boldsymbol{ {\epsilon}}_{\pi}$ to 100. These two configurations yield better results than (a) and (b). This is because the target distribution of (c) and (d) is a 3D head-aware diffusion model. Nonetheless, due to the spatial mismatch of noise during gradient calculation of (c), the predicted results still exhibit the multi-face Janus issue. And (d)  suffers from over-saturation and over-smoothing with large CFG, which are the inherent problems of SDS. In contrast, our full configuration (e) achieves better performance in both geometry (i.e., able to generate dreadlocks) and texture with SSD, which leverages head prior knowledge effectively and benefits from the spatial alignment of the predicted scores from two identical ControlNets. We also provide the corresponding quantitative results in Table.~\ref{tab:ablation}.

\subsubsection{ Effectiveness of Negative Prompts}
We observe that our method can directly employ negative prompts to improve further performance. Without the use of negative prompts, as shown in (f) of Figure~\ref{fig:ablation}, our method tends to exhibit slight over-exposure and may introduce artifacts (i.e., there are some texts printed on the face of the first man). This artifact is also observed in the 2D images generated by ControlNet if negative prompts are not employed. By using negative prompts, we have improved the quality of the generated content while effectively reducing artifacts. We also provide the corresponding quantitative results in Table.~\ref{tab:ablation} of supplementary.

\section{Limitations}
 \begin{figure}[t]
\begin{center}
\includegraphics[width=0.8\linewidth]{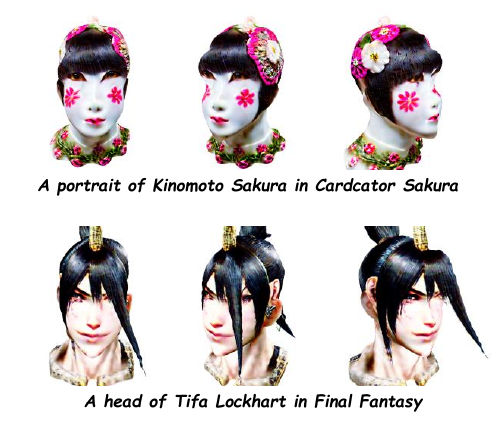}
\caption{The failure results of our method. We find our method can not handle such complex Japanese animation characters well.}
\vspace{-0.25in}
\label{fig:limitation}
\end{center}
\end{figure}
 Our method currently cannot achieve the photorealism of 3D head reconstruction methods or 3D GANs. Additionally, it is difficult to generate some characters with complex structures in Japanese animation (e.g., KINOMOTO SAKURA in Fig.~\ref{fig:limitation}). This is because FLAME initializes our model, and no geometric priors of Japanese anime characters are introduced. Moreover, the diffusion model we used is not good at generating Japanese animation characters. Meanwhile, our method can change the expression of the character, but we cannot drive the 3D head directly, which is also one of our future works.

\section{Conclusion}
We propose HeadArtist with Self Score Distillation (SSD) to generate the high-fidelity 3D human heads with language expressions. Our SSD utilizes a landmark-guided ControlNet twice to predict the distribution scores of the optimized and target heads, respectively. Then, we minimize the distance of these two scores to update the parameters of the 3d head. The two scores are spatially aligned to avoid the multi-face Janus issue. Meanwhile, the landmark introduces facial semantics. Moreover, we can utilize the negative prompts to enhance the texture quality. Experiments on diverse text prompts demonstrate the significant performance of our method. Finally, we extend the SSD to general object generation by replacing the ControlNet with other 3D-aware diffusion models (i.e., MVDream~\cite{liu2023zero}) in supplementary. This attempt could bring a new perspective to 3D generation.

 \begin{acks}
This project was supported by the National Key R\&D Program of China under grant number 2022ZD0161501.

\end{acks}
 
\clearpage

\bibliographystyle{ACM-Reference-Format}
\bibliography{sample-base}
 \begin{figure*}[t]
\begin{center}
\includegraphics[width=1\linewidth]{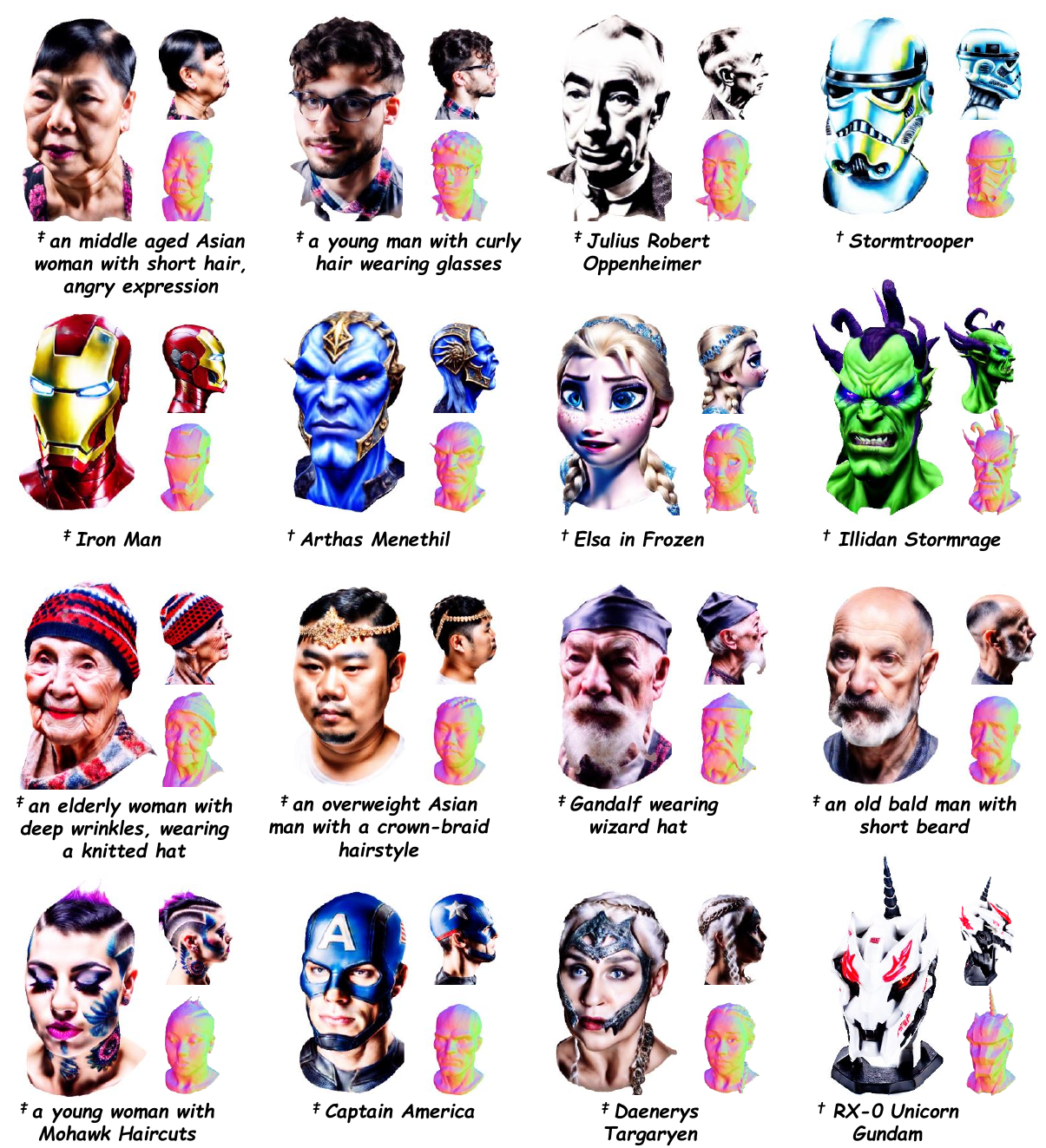}
\caption{More generation results of HeadArtist. Our method preserves remarkable performance on both authentic human and cartoon character heads generation. $\dag$ and $\ddag$ denote the prefixes ``a head of ...'' and ``a DSLR portrait of ...'', respectively.}
\vspace{-0.23in}
\label{fig:gallery}
\end{center}
\end{figure*}

 \begin{figure*}[t]
\begin{center}
\includegraphics[width=0.76\linewidth]{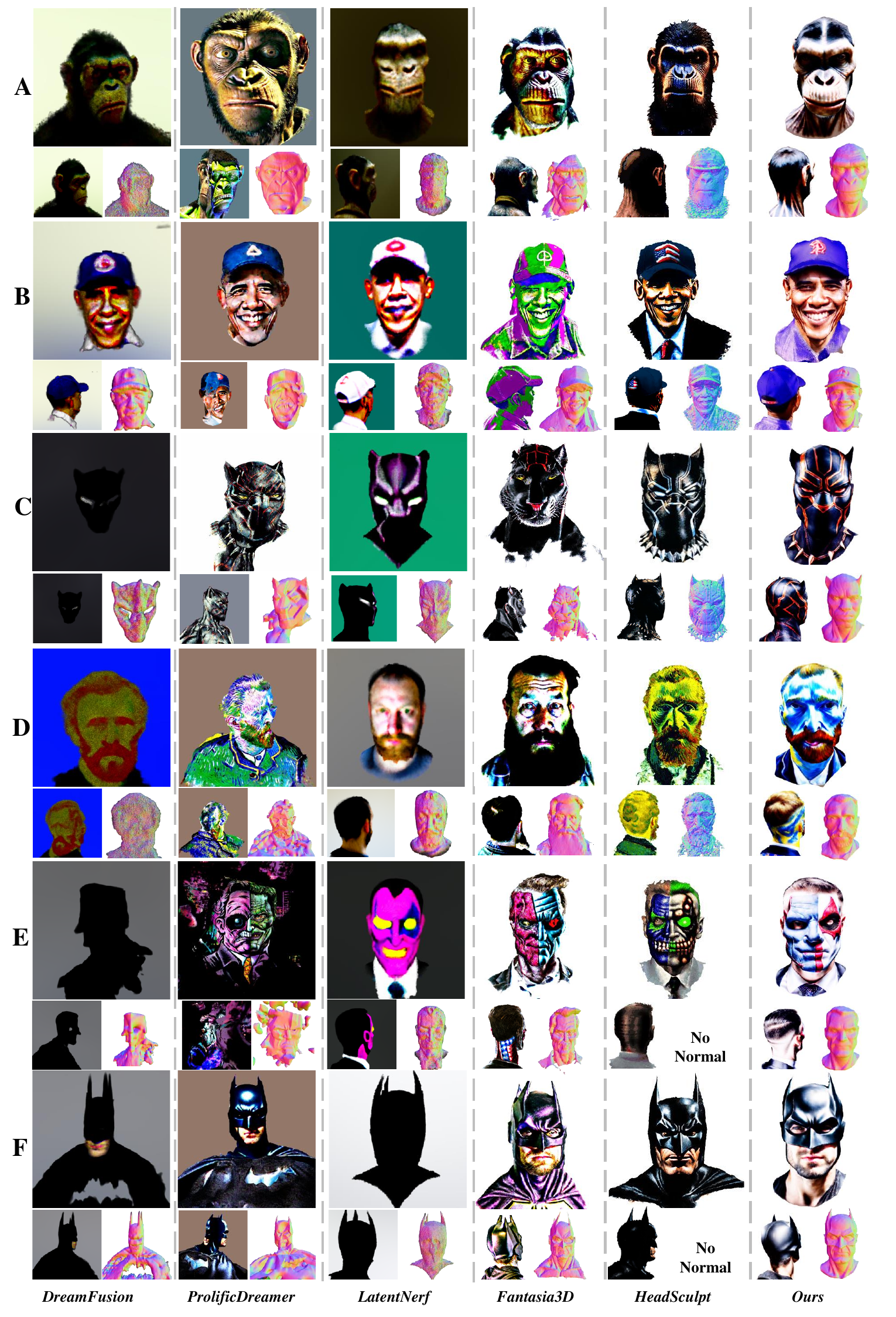}
\caption{ More qualitative comparison. Text Prompts: (A) a DSLR portrait of Caesar in Rise of the Planet of the Apes. (B) a DSLR portrait of Obama with a baseball cap. (C) a DSLR portrait of Black Panther in Marvel. (D) a portrait of Vincent Van Gogh. (E)  a DSLR portrait of Two-Face in DC. (F) a DSLR portrait of Batman. Visually, our method generates more reasonable geometry and high-quality texture, and the generated 3D heads are more consistent with the corresponding text prompts. }
\vspace{-0.23in}
\label{fig:compare_supp_1}
\end{center}
\end{figure*}

\appendix
\clearpage
\setcounter{page}{1}

In this supplementary, we first show that our method can effectively address the Janus problem in general object generation. Then, we provide visual comparison results between RichDreamer, LucidDreamer, HumanNorm, Magic123, and Our method. The results demonstrate the excellent performance of our method. Furthermore, we discuss the limitations of our method and display the result when our method abandons the texture step or utilizes other 3D representations. Meanwhile, we provide more details of our method, including pseudocode, 3D representation initialization, and the weights for each loss item.  Finally, we show diverse results when we use the same prompt as guidance.


 \begin{figure}[t]
\includegraphics[width=0.5\linewidth]{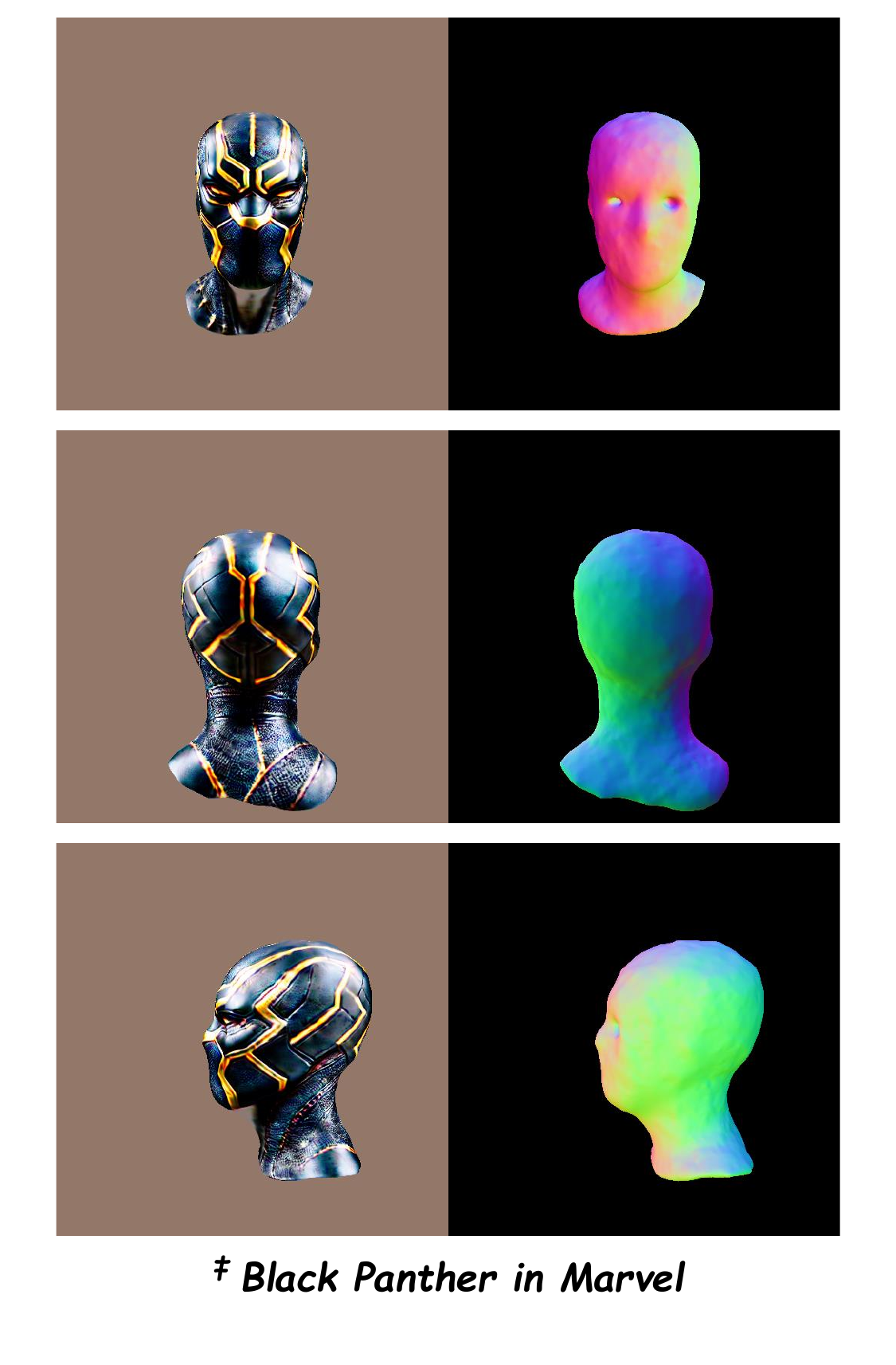}
\caption{The generation results when HeadArtist merges the texture and geometry steps. We find the geometry is hard to optimize. During the generation, we set the CFG to 7.5 to avoid over-saturation and over-smooth, but geometry needs a large CFG for stable training since there is a domain gap between the real image and the normal map.}

\label{fig:Notexture}
\end{figure}

 \begin{figure}[t]
\includegraphics[width=1.\linewidth]{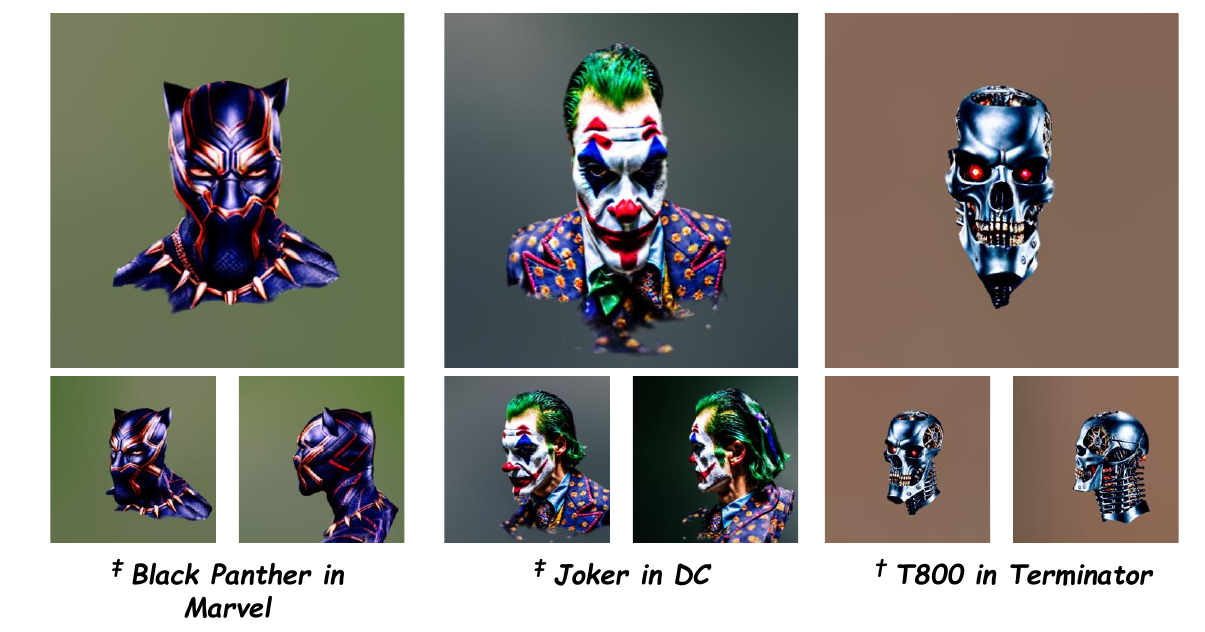}
\caption{The generation results when HeadArtist utilizes NeRF representation. The good performance shows our method can be adapted to different representations. }

\label{fig:nerf}
\end{figure}

 \begin{figure*}[t]
\begin{center}
\includegraphics[width=1\linewidth]{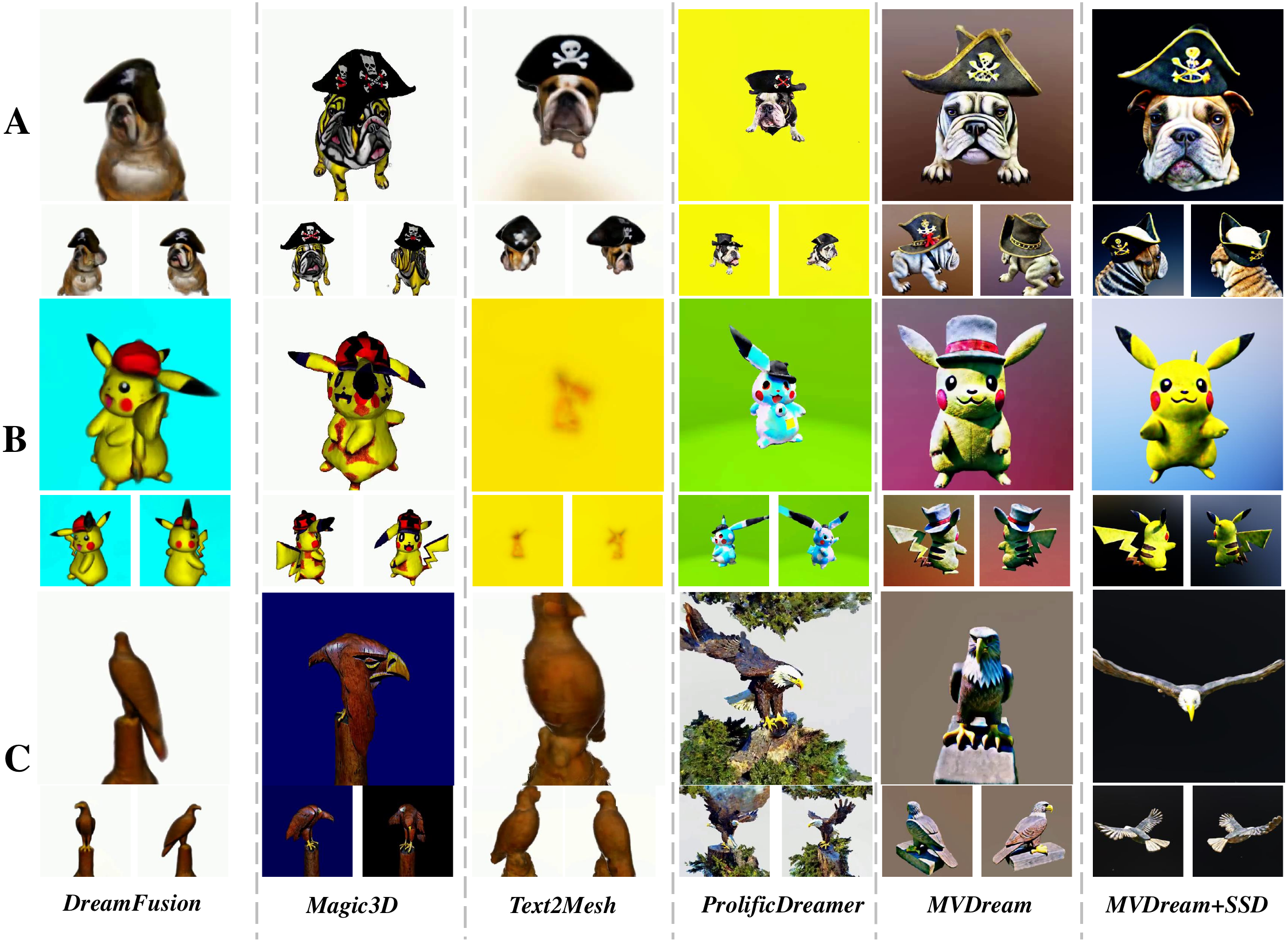}
\caption{ Text Prompts: (A) A bulldog wearing a black pirate hat. (B) A bald eagle carved out of wood. (C) Pikachu with hat. Overall, MVDream+SSD demonstrates superior fidelity in terms of both geometry and textures.}
\vspace{-0.23in}
\label{fig:Janusproblem}
\end{center}
\end{figure*}

\section{Resolving the Janus problem in general object generation with SSD}
We replace the Controlnet with MVDream and the DMTed with Nerf for general object generation. Meanwhile, the geometry and texture are generated together in a single step, similar to the DreamFusion. And we initialize the NerF with spatial density bias following the Magic3D.  For fair comparisons, we use the results on the project page of MVDream for all comparable methods (i.e., DreamFusion, Magic3D, Text2Mesh, ProlificDreamer, and MVDream). The results are shown in Fig.~\ref{fig:Janusproblem}. Apart from the MVDream and MVDream+SSD, the other methods all have the Janus problem. This is because the MVDream is a multi-view diffusion model, which brings the pose information during the generation process to alleviate the Janus problem. However, MVDream utilizes the SDS directly for 3D asset generation in the original paper, which causes the results to be unreal and over-saturation (e.g., the unnatural texture in b of MVDream). In contrast, we can produce more reasonable textures and address the Janus problem at the same time, which proves the advantages of our approach (e.g., more real results in b of MVDream+SSD ).

 \begin{figure*}[t]
\begin{center}
\includegraphics[width=1\linewidth]{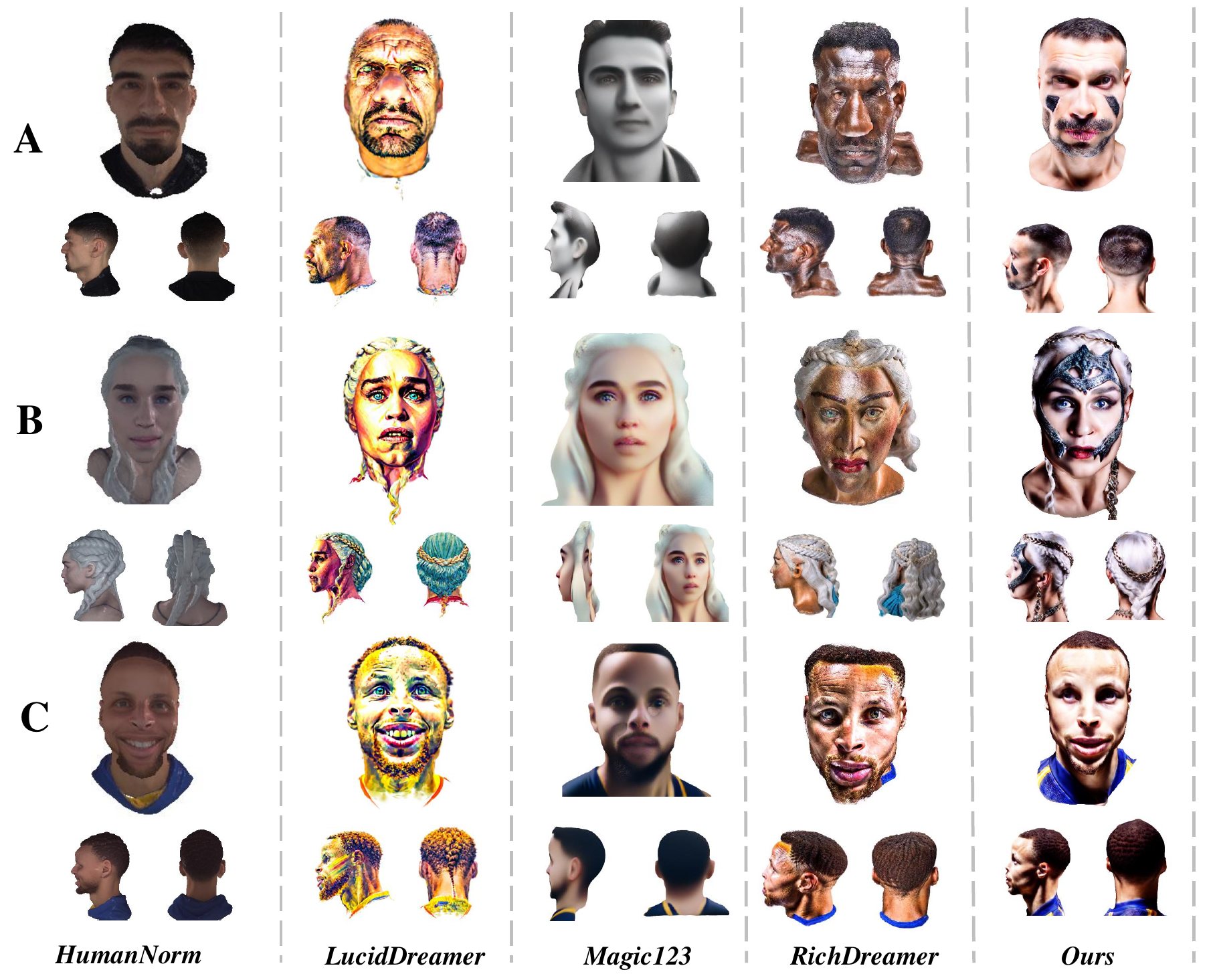}
\caption{ Text Prompts: (A) a DLSR portrait of a man with a square jaw. (B) a DSLR photo of Leonardo DiCaprio (C) a DSLR photo of Stephen Curry. Overall, our method demonstrates superior fidelity in terms of both geometry and textures.}
\vspace{-0.23in}
\label{fig:MoreCompare}
\end{center}
\end{figure*}

\section{Visual comparison with more methods}
We compare our method with RichDreamer, LucidDreamer, HumanNorm, and Magic123 in Fig.~\cite{fig:MoreCompare}. The HumanNorm can produce good results, but the content is not real (i.e., not fidelity texture in b of HumanNorm) and the geometry seems distorted (i.e., exaggerated geometry in c of HumanNorm). The LucidDreamer generates the results with both strange colors and over-saturation. For Magic123, we generate the input image with ControlNet and the corresponding text prompt. We find the results of  Magic123 suffers from the Janus problem and over-smooth. RichDreamer performs well in c, but the geometry has artifacts (i.e., the head of c in RichDreamer is not completed ). For the a and b, RichDreamer shows bad performance (i.e., the artifacts in both eyes in a and the unreal content in b). In contrast to these baselines, our method generates human heads with more fidelity texture and reasonable geometry simultaneously.

\section{Necessity of texture step and utilizing other 3D representation}
We merge the texture and geometry generation in one step, as shown in Fig.~\ref{fig:Notexture}. We find our method can produce good texture but the geometry is hard to optimize (e.g., the geometry is similar to the shape of Flame). To merge the two steps more efficiently, we replace the DMTed with NeRF for our head generation. The results are shown in Fig.~\ref{fig:nerf}, and our HeadArtist can provide plausible geometry and fine detail texture.


\section{More details}
We describe the pseudocode in Algorithm~\ref{alg:algorithm1}. The min\_step and max\_step are the maximize and minimize noise levels, respectively. For the first 5000 iterations, the max\_step is 980, then it will decrease to 700 to make the training focus on detail generation similar to the ProlificDreamer. During the geometry generation, we use two common regularization losses, including normal consistency and Laplacian smoothness. The weights of these two losses are 8000 and 10000, respectively. For the geometry initialization, we randomly select 10000 points for each iteration, and we minimize the distance between the sdf value of these points from the Flame model and optimize DMTed.  For the texture,  we set the maximum time steps as 0.98, and we decrease it to 0.7 after 5000 iterations for detail generation similar to ProlificDreamer~\cite{wang2023prolificdreamer}. For the camera sampling, we set the camera distance range as 3, set the FOV range as $(30, 50)$, and set the elevation range as $(-10, 45)$.

Finally, we follow HIFA~\cite{zhu2023hifa} to calculate the SSD in image space with around 900 iterations after the texture generation to get better performance.

\begin{algorithm}[t]
	\caption{Self Score Distillation.}
	\label{alg:algorithm1}
	{\alliu{def:}} self score distillation(landmark $c_L$, ControlNet, rendered\_image, text $y$, negative\_text $y_{neg}$): \\
    \begin{algorithmic}
        \STATE $\text{t} = \text{torch.randint(min\_step, max\_step)}$
        \STATE $\text{noise} = \text{torch.randn\_like(rendered\_image)}$
        \STATE $\boldsymbol{x}_t = \text{rendered\_image} + \text{noise}$
        \STATE $\text{noise}_{\text{pretrain}} = \text{Controlnet}(\boldsymbol{x}_t ; y, y_{neg}, t, c_L, CFG=7.5 or 100)$
        \STATE $\text{noise}_{\text{head}} = \text{Controlnet}(\boldsymbol{x}_t ; y, t, c_L, CFG=1)$
        \STATE $\text{grade} = \text{noise}_{\text{pretrain}} - \text{noise}_{\text{head}}$
        \STATE target = (rendered\_image - grad).detach()
        \STATE loss = MSE(rendered\_image, target) 
    \end{algorithmic}

\end{algorithm}

\section{Diverse results}
\begin{figure*}[t]
\begin{center}
\includegraphics[width=1\linewidth]{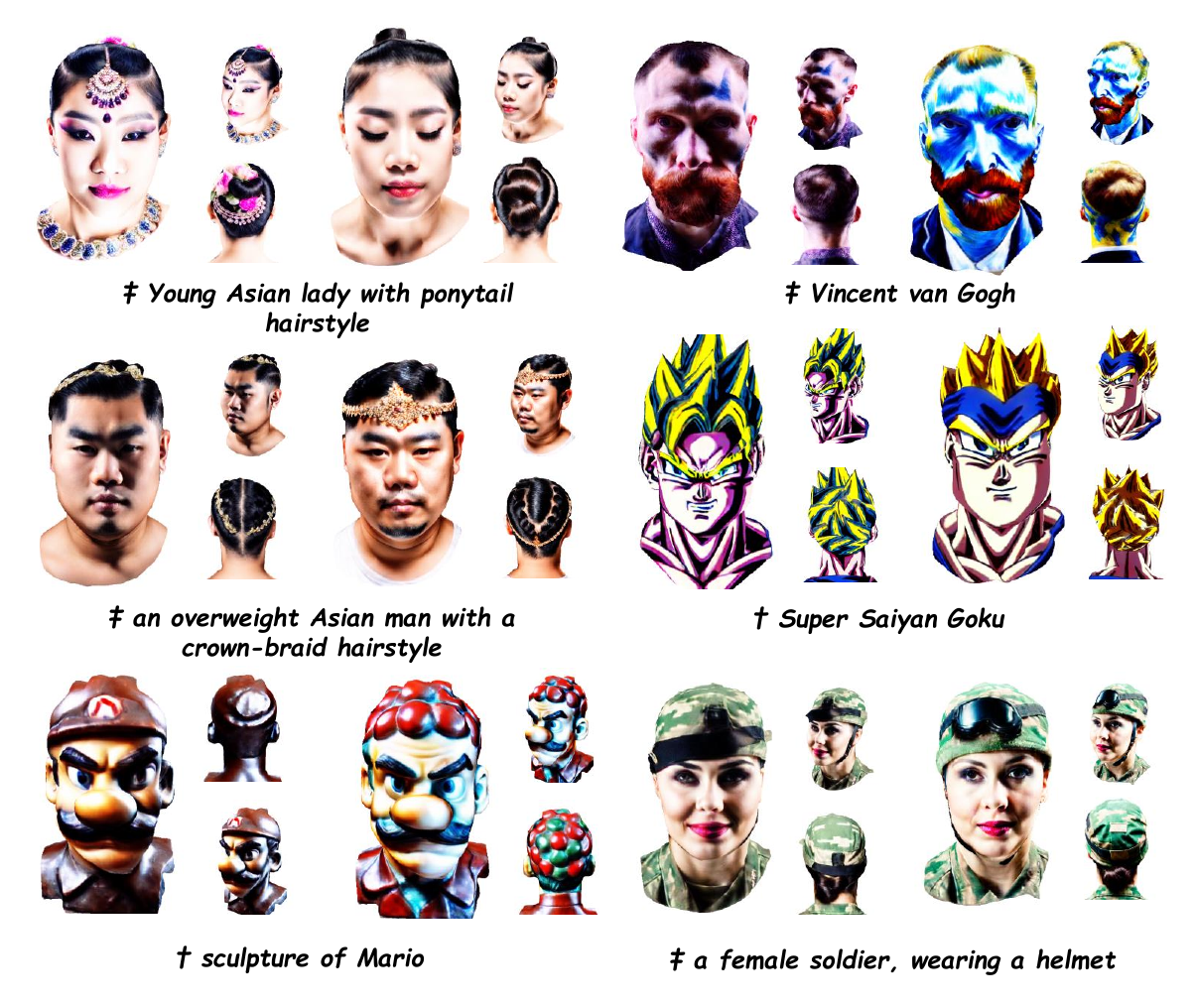}
\caption{The diverse results of our method. Our method can produce different high quality results with a single text prompt. $\dag$ and $\ddag$ denote the prefixes ``a head of ...'' and ``a DSLR portrait of ...'', respectively.}
\vspace{-0.23in}
\label{fig:diversity}
\end{center}
\end{figure*}

In Fig.~\ref{fig:diversity}, we show our method can generate diverse results with the same promotes, which demonstrates the robustness of our method.

\section{More editing results}
\begin{figure*}[t]
\begin{center}
\includegraphics[width=1\linewidth]{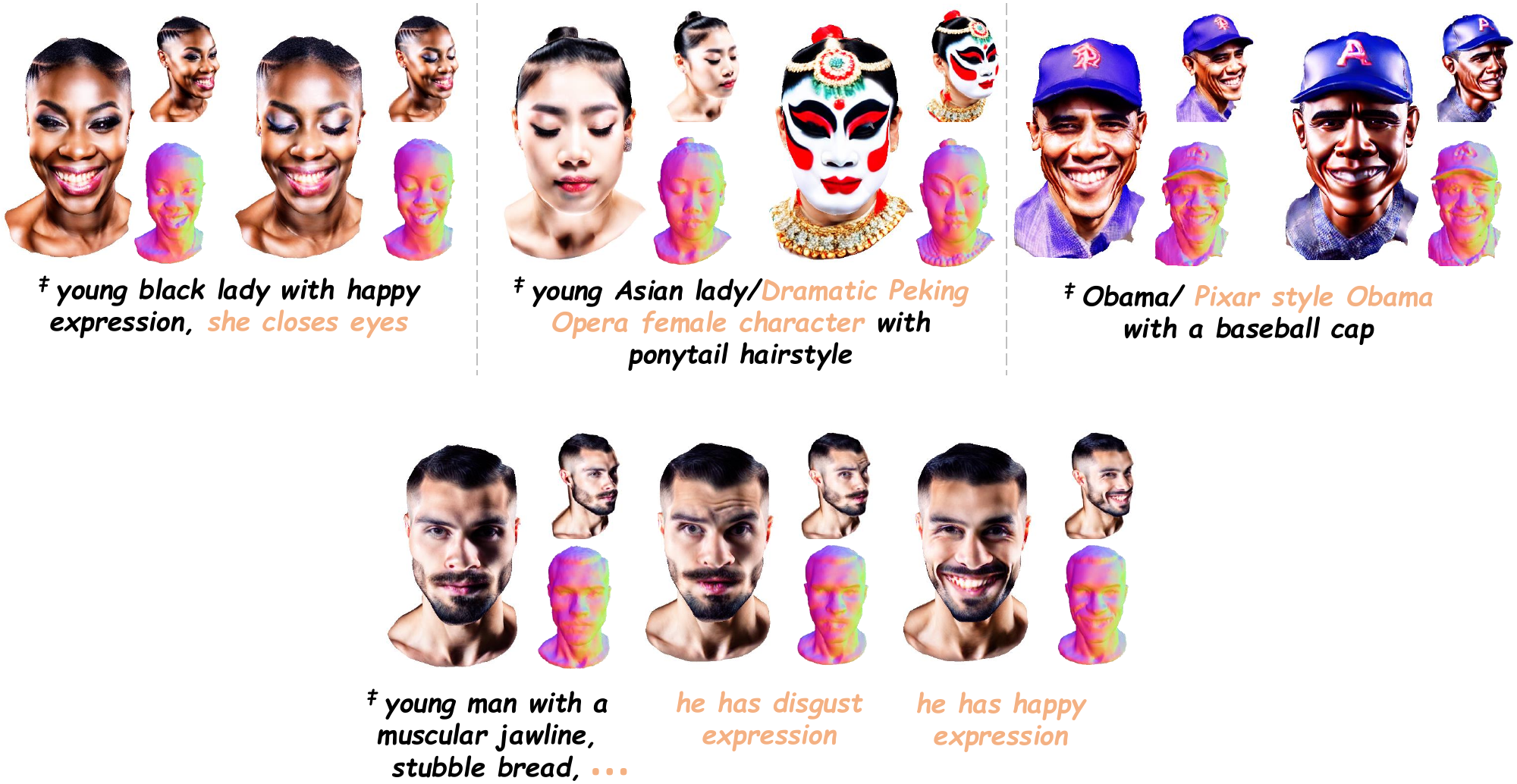}
\caption{More editing results of HeadArtist. Our method can effectively modify the texture and geometry to match the corresponding text semantics. Text in \textbf{\textcolor[rgb]{0.95,0.69,0.51}{orange}} denotes the editing instruction.}
\vspace{-0.23in}
\label{fig:editing_supp}
\end{center}
\end{figure*}

Figure~\ref{fig:editing_supp}  demonstrates our ability to edit  3D head models in terms of geometry and texture effectively.

\end{document}